\pdfoutput=1
\documentclass[11pt]{article}

\usepackage[dvipsnames]{xcolor}
\usepackage{naacl2021}

\usepackage[utf8]{inputenc}

\usepackage{times}
\usepackage{latexsym}
\usepackage{xspace}
\usepackage{enumitem}

\usepackage{url}

\usepackage{booktabs}
\usepackage{diagbox}
\usepackage{multirow}
\usepackage{balance}
\usepackage{amsfonts}
\usepackage{amsmath}
\usepackage{amssymb}
\usepackage{graphicx}
\usepackage{subfigure}
\usepackage{microtype}
\usepackage{wrapfig,lipsum}
\usepackage{algorithm}
\usepackage{algorithmicx}
\usepackage{algpseudocode}
\usepackage{makecell}
\usepackage[T1]{fontenc}
\usepackage{soul}
\usepackage{tabularx, booktabs}
\usepackage{todonotes}
\usepackage{comment}

\newcommand{\cbb}{\mathbf{c}}
\newcolumntype{Y}{>{\centering\arraybackslash}X}
\newcolumntype{L}{>{\arraybackslash}X}

\algnewcommand\algorithmicinput{\textbf{Input:}}
\algnewcommand\algorithmicoutput{\textbf{Output:}}
\algnewcommand\algorithmicparameter{\textbf{Parameters:}}
\algnewcommand\INPUT{\item[\algorithmicinput]}
\algnewcommand\OUTPUT{\item[\algorithmicoutput]}
\algnewcommand\PARAMETER{\item[\algorithmicparameter]}

\newcommand{\clstoken}{\texttt{[CLS]}}
\newcommand{\septoken}{\texttt{[SEP]}}
\newcommand{\kldiv}{\text{KL}}

\newcommand{\eat}[1]{\ignorespaces}
\input{Definitions.tex}

\title{
Posterior Differential Regularization with $f$-divergence for Improving Model Robustness
}

\author{
Hao Cheng\textsuperscript{1}, Xiaodong Liu\textsuperscript{1}, Lis Pereira\textsuperscript{2}, Yaoliang Yu\textsuperscript{3}, Jianfeng Gao\textsuperscript{1}
 \\ 
  \textsuperscript{1}Microsoft Research 
  \textsuperscript{2}Ochanomizu University
  \textsuperscript{3}University of Waterloo \& Vector Institute
 \\
  {\tt \{chehao,xiaodl,jfgao\}@microsoft.com} \\
  {\tt \{kanashiro.pereira\}@ocha.ac.jp} \\
  {\tt \{yaoliang.yu\}@uwaterloo.ca} 

}

\date{}

\begin{document}
\maketitle

\begin{abstract}
We address the problem of enhancing model robustness through regularization.
Specifically, we focus on methods that regularize the model posterior difference between clean and noisy inputs.
Theoretically, we provide a connection of two recent methods, Jacobian Regularization and Virtual Adversarial Training, under this framework.
Additionally, we generalize the posterior differential regularization to the family of $f$-divergences and characterize the overall framework in terms of Jacobian matrix.
Empirically, we compare those regularizations and standard BERT training on a diverse set of tasks to provide a comprehensive profile of their effect on model generalization.
For both fully supervised and semi-supervised settings, we show that regularizing the posterior difference with $f$-divergence can result in well-improved model robustness.
In particular, with a proper $f$-divergence, a BERT-base model can achieve comparable generalization as its BERT-large counterpart for in-domain, adversarial and domain shift scenarios, indicating the great potential of the proposed framework for enhancing NLP model robustness.\footnote{Code is available at \url{https://github.com/hao-cheng/f-divergence}.}

\end{abstract}

\section{Introduction}
\label{sec:intro}

\begin{figure}[t]
    \centering
    \includegraphics[width=0.48\textwidth]{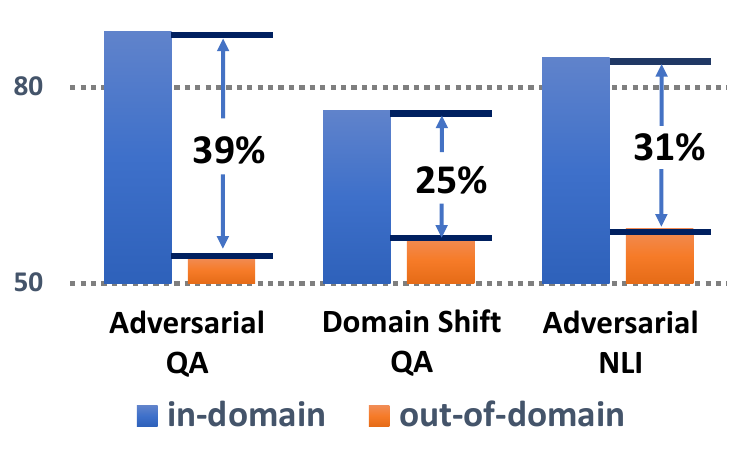}
    \caption{
    BERT-based question answering (QA) and natural language inference (NLI) model robustness towards domain shift and adversarial attack. The numbers in percentage indicate the corresponding relative performance drop between in-domain and out-of-domain test sets.}
    \label{fig:bert_robustness}
\end{figure}

Although recent neural network based models have achieved great success in a wide range of natural language processing (NLP) tasks, these models may still suffer catastrophic degradation in out-of-domain generalization to datasets with domain shift or adversarial scenarios \cite{nie2019adversarial, hsieh2019robust-self-att}. 
For example, large-scale pretrained neural language models \cite{devlin2018bert,liu2019roberta} have better generalization,
but experience performance reduction caused by domain shift \cite{hendrycks2020acl}.
Textual entailment models trained on MNLI \cite{mnli2018} have picked up superficial cues focusing on either the presence of certain keywords \cite{gururangan2018annotation} or whether similar words are mentioned in the sentence pairs \cite{mccoy2019acl}. 
\citet{jia2017advsquad} have also shown that SQuAD models are very easily distracted by irrelevant sentences that contain many question words regardless of context, despite of
their human-performance on in-domain data.
As shown in Figure~\ref{fig:bert_robustness}, three BERT-based \cite{devlin2018bert} models perform well for in-domain evaluation data, but transfer poorly to out-of-domain datasets with domain shift or adversarial attack, \ie more than $25\%$ relative performance reduction.

Achieving good generalizations towards datasets with domain shift has been a long-standing goal of domain adaptation.
Various methods \cite{blitzer-etal-2007-biographies,daume-iii-2007-frustratingly} have been developed for training models to learn effectively from both in-domain (source) and out-of-domain (target) datasets.
Additionally, the recent discovery on the prevalence of data \textit{biases} \cite{gururangan2018annotation,jia2017advsquad,mccoy2019acl}, unintended correlations between input and output learned by statistical models, ignites the development of model debiasing techniques \cite{clark2019emnlp,he2019unlearn}.
These methods leverage discovered data biases to improve model generalization over adversarial datasets \cite{jia2017advsquad,mccoy2019acl} designed to fool naively trained models. 
Instead of relying on knowledge of the target dataset, we focus on task-agnostic training techniques for enhancing model robustness with access to only in-domain data.

Motivated by recent success of adversarial training in computer vision \cite{madry2017pgd,goodfellow2014explaining} and NLP \cite{zhu2019freelb,jiang2019smart,cheng-etal-2019-adv-nmt,wang2019adv-lm},
we investigate a regularization framework, which directly regularizes the model posterior difference for clean and noisy inputs, as a means to enhance the model robustness.
Here, we first provide a theoretical connection of two recent methods under this framework, \ie Virtual Adversarial Training (VAT) \cite{miyato2018virtual} and Jacobian Regularization (JR) \cite{sokilic2017TSP}.
In addition, we propose to generalize VAT and random perturbation training (RPT) \cite{miyato2018virtual} with a family of probability distribution metrics, $f$-divergences, and characterize their connection with JR. 

Given that large-scale pretrained neural language models have demonstrated their superior generalization for downstream NLP tasks under both matched \cite{devlin2018bert,liu2019roberta} and mismatched evaluations \cite{hendrycks2020acl}, we systematically study the regularization framework using BERT \cite{devlin2018bert} on a diverse set of tasks in terms of both in-domain and out-of-domain generalization.
Specifically, we use representative datasets \cite{sst2013,mnli2018,squad1,squad2} from sentiment analysis, textual entailment and question answering (QA) for in-domain training and evaluation.
In order to assess the resulting model generalization over domain shift and adversarial attack, we then consider out-of-domain datasets \cite{bioasq,imdb2011} and challenge adversarial datasets \cite{jia2017advsquad,mccoy2019acl} in a zero-shot learning fashion.

Our experiments show that regularizing the posterior difference for clean and noise inputs is very effective in improving model generalization under both supervised and semi-supervised learning settings.
Based on our theoretical analysis, both VAT and RPT variants, unlike JR, incorporate model confidence for adaptive regularization, which leads to consistently better empirical robustness over BERT with the standard fine-tuning.
Furthermore, we find that different $f$-divergences lead to different generalization behaviors for in-domain, domain shift and adversarial settings.
In our study, VAT with symmetric divergence achieve better generalization for in-domain and domain shift cases, while
VAT with asymmetric divergence achieve more robustness toward adversarial attack.
More importantly, we show that a BERT-base model trained with a proper $f$-divergence can perform comparably to its corresponding BERT-large counterpart. It is also worth noting that VAT with symmetric divergence lead to improved data efficiency, \ie achieving comparable in-domain performance as fully-supervised models with only 50$\%$ labelled data.
This further illustrates the great potential of the proposed general regularization framework for the semi-supervised setting.

Our main contributions are summarized as follows: 
1) we generalize the posterior differential regularization framework to the family of $f$-divergences and provide additional divergence functions with different characteristics for regularization;
2) based on our framework, we analyze the family of regularization methods and show the theoretical connection of recently proposed methods, JR, VAT and RPT, in terms of their regularization effect on the input-output Jacobian matrix;
3) We provide a comprehensive profile of different regularization methods over a diverse set of NLP tasks and experimental insight into which $f$-divergence is more suitable for improving NLP model robustness under both supervised and semi-supervised settings. 
\section{Posterior Differential Regularization}
\label{sec:robust_learning}


In this section, we first introduce the regularization framework that penalizes the difference of posterior between clean and noisy inputs.
Based on this framework, we set up the basic notions of two recent methods, \ie VAT and JR, and show their theoretical connection.
Finally, we generalize the posterior differential regularization with any function from the family of $f$-divergences and characterize their local smoothness promoting in terms of Jacobian matrix.
In the following, we use $f_\Theta(\xvec):\RR^n\to\RR^m$ to denote the posterior function
which is a neural network parameterized by $\Theta$ that maps the input $\xvec\in\RR^n$ to the output probability space with $m$ discrete classes.

Both adversarial learning \cite{goodfellow2014explaining} and the posterior difference regularization aim at making the model more robust.
Adversarial learning focuses on minimizing the following objective
\begin{eqnarray}
\min_\Theta \max_{\|\epsilon\|\leq c} L(f_\Theta(\xvec+\epsilon), y),
\end{eqnarray}
where $L$ is the cross-entropy loss, $\epsilon\in\RR^n$ is a random vector bounded in a norm by $c$, a small positive constant and $\yvec$ is the target label.
Instead, the posterior differential regularization directly promotes the model local smoothness, \eg stabilizing the model posterior distribution towards small input perturbations.
Typically, it is in the form of 
\begin{eqnarray}
\min_\Theta L(f_\Theta(\xvec), \yvec) + \alpha R(f_\Theta(\xvec), f_\Theta(\hat{\xvec})),
\end{eqnarray}
where $\hat{\xvec} = \xvec + \epsilon$, $R$ is a regularization term penalizing the model instability, and $\alpha$ is a hyperparameter for balancing the classification loss and the regularization term.
As we can see, posterior differential regularization is a task-agnostic method which makes it applicable to semi-supervised, self-supervised and unsupervised learning.
For simplicity, we will use $f$ and $R$ to denote $f_\Theta$ and $R(f_\Theta(\xvec), f_\Theta(\hat{\xvec}))$, respectively.

\label{ssec:jacobian}
\noindent\textbf{Jacobian Regularization:} A recent regularization approach to stabilize the model is Jacobian regularization \cite{sokilic2017TSP,li-etal-2016-text-jr}.
Specifically, using the input-output Jacobian matrix, $J=J_f(\xvec)\in\RR^{m\times n}$, 
we can get the first-order Taylor approximation
\begin{equation}
    f(\hat{\xvec}) = f(\xvec + \epsilon) = f(\xvec) + J_f(\xvec)\epsilon.
\end{equation}
In order to reduce the overall model sensitivity to the input perturbation, \citet{sokilic2017TSP} propose to directly regularize the Frobenius norm of the input-output Jacobian matrix so that
\begin{eqnarray}
\label{eqn:jacobian_norm}
\nonumber
\|f(\hat{\xvec}) - f(\xvec)\|_2^2 & = & \|J\epsilon\|_2^2=\epsilon^TJ^TJ\epsilon \\
\nonumber
& \leq& \|\epsilon\|_2^2 \|J\|_{sp}^2 
\leq \|\epsilon\|_2^2 \|J\|_{F}^2,
\end{eqnarray}
where $\|\cdot\|_2$ is the $L2$ norm, $\|\cdot\|_{sp}$ is the spectral norm, and $\|J\|^2_F=\tr(J^TJ)$ is the Frobeinus norm of the Jacobian matrix with $\tr$ as the trace operator.
In other words, by letting $R = \|J\|^2_F$, the $L2$ difference between clean and noisy inputs is thus being effectively regularized.

\noindent
\textbf{Virtual Adversarial Training:}
Motivated by the adversarial learning objective used in \cite{goodfellow2014explaining}, \citet{miyato2018virtual} introduce a regularized objective to enhance the model robustness towards small input perturbations 
\begin{align}
\label{eqn:vat}
    \min [L(\yvec, \hat{\yvec}) + \alpha\underbrace{\max_{\|\epsilon\|\leq c} \kldiv(\hat{\yvec}, f(\xvec+\epsilon))}_{R}],
\end{align}
where $\kldiv$ is the well-known \textit{Kullback-Leibler divergence} , and $\hat{\yvec}=f(\xvec)$. 
Based on the above definition, VAT essentially regularizes the KL-based worst-case posterior difference between the clean and noisy input using an inner loop to search for the most adversarial direction.

Although sharing with JR the same spirit of encouraging the local smoothness of the model, JR and VAT are not fully theoretically connected.
In what follows, we use a simple approach to draw the theoretical connection between these two methods.

\noindent\textbf{Connection between VAT and JR:}
Here, we show that VAT and JR can be directly related through the definition of induced matrix norm.
Specifically, the matrix norm of the Jacobian matrix is
\begin{eqnarray}
\|J\| = \sup_{\|\nu\|=1} \|J\nu\|,
\end{eqnarray}
where the matrix norm on the left side is induced by the corresponding vector norm on the right side.
It is easy to show that
\begin{eqnarray*}
c^2\|J\|_{sp}^2 & \approx & \sup_{\|\epsilon\|_2=c} \|J\epsilon\|_2^2 \\
& \leq & \sup_{\|\epsilon\|_2=c} \|f(\hat{\xvec})-f(\xvec)\|_1^2 \\
& \leq & 2\sup_{\|\epsilon\|_2=c} \kldiv(f(\xvec), f(\hat{\xvec})),
\end{eqnarray*}
and the last inequality is attained based on Pinsker's inequality.
Therefore, the VAT regularization provides an upper bound for the spectral norm of the Jacobian matrix.
Although a similar attempt to relate JR and VAT has been first explored in \cite{Abbas2016UnderstandingRB}, we provide a simple and comprehensive connection. Specifically, both VAT and JR regularize the upper bound of the spectral norm of the Jacobian matrix.

\label{ssec:pdr}
\noindent\textbf{Posterior Differential Regularization with $f$-divergence:}
Although both VAT and JR have been successful in improving model robustness, they are both special cases of regularizing the model posterior difference between the clean and noisy inputs. 
One natural question is whether we can use other probability distribution metrics for regularization and characterize them in terms of Jacobian matrix.
In the following, we extend the posterior difference regularization with the family of $f$-divergences \cite{f-divergence}.
Furthermore, we show that posterior differential regularization with all $f$-divergences results in an adaptive variant of JR which incorporates model confidence.

First, let's define the \textit{$f$-divergence} for measuring the posterior difference between the clean and noisy inputs, \ie 
\begin{equation}
    D_g\left( f(\hat{\xvec}), f(\xvec)\right) = 
    \sum_i f_i(\xvec) g\left( \frac{f_i(\hat{\xvec}) }{f_i(\xvec)}\right),
    \label{eqn:fdiv}
\end{equation}
where the \textit{generator function} $g: \RR_+ \mapsto \RR$ is a convex and lower-semicontinuous function satisfying $g(1) = 0$, $\hat{\xvec} = \xvec + \epsilon$ and $f_i$ indicates the $i$-th element of vector $f$.
Different choices of $g$ lead to several popular divergences, \eg KL, squared Hellinger and Jensen-Shannon divergence.
Based on this, it is easy to show that the corresponding second order approximation is
\begin{eqnarray}
    D_g\left( f(\hat{\xvec}), f(\xvec)\right) \approx {g^{''}(1) \over 2}
     \epsilon^T J^T\diag{1\over f} J \epsilon,
    \label{eqn:fdiv_approx_new}
\end{eqnarray}
where $J$ is the input-output Jacobian of $f$, and $\diag{1\over f}$ is a diagonal matrix with elements equal to ${1\over f}$ (See \autoref{apd_sec:approx_reg_derive} for full derivation).

Compared with the Frobenius norm of Jacobian matrix $\|J\|_F^2$, \autoref{eqn:fdiv_approx_new} can be seen as a weighted version of JR where each row is rescaled by the model confidence $f_i$ for the corresponding class.
In other words, it is close to JR for more confident classes,
whereas for uncertain classes it allows less Jacobian variance.
Additionally, although $g^{''}(1)$ is a constant once the generator function is selected, various $f$-divergences can lead to different approximations which might result in task-dependent benefits.
Therefore, different from KL-based VAT or its sampling alternative without the inner search for the most adversarial direction as proposed in \cite{miyato2018virtual}, we generalize the posterior differential regularization with the family of $f$-divergences and show that they all provide an approximation to a variant of JR which adapts the regularization based on model confidence.

\eat{
Based on \autoref{eqn:fdiv}, it is easy to show that the first and second derivatives of $D$ with regard to $\epsilon$ are
\begin{eqnarray*}
    D^\prime & = & \sum_i
    g^\prime(\rvec_i)
    f_i^\prime(\hat{\xvec}), \\
    D^{\prime\prime} & = & \sum_i\left[
     g^{\prime\prime}(\rvec_i)   
    {f^{\prime}_i(\hat{\xvec}){f^\prime_i}^T(\hat{\xvec}) \over f_i(\xvec)}
    + 
    g^\prime(\rvec_i)
    f_i^{\prime\prime}(\hat{\xvec})
    \right],
\end{eqnarray*}
where $\rvec_i = {f_i(\hat{\xvec}) \over f_i(\xvec)}$.
Let $D(\epsilon) = D(f(\hat{\xvec}), f(\xvec))$.
When evaluating at $0$, the second order approximation of the $f$-divergence is
\begin{eqnarray}
    D(\epsilon) \approx D(0) + {D^{\prime}}^T(0)\epsilon + {\epsilon^T D^{\prime\prime}(0) \epsilon \over 2}.
    \label{eqn:fdiv_2nd_approx}
\end{eqnarray}
Given that 
\begin{eqnarray*}
D(0) & = & \sum_i g(1)f_i = 0, \\
D^\prime(0) & = & \sum_i g^\prime(1)f_i^\prime = g^\prime(1)
\left(\sum_i f_i\right)^\prime = 0, \\
\sum_i f^{''}_i & = & \left(\sum_i f_i\right)^{''} = 0
\end{eqnarray*}
the second order approximation can be simplified as
\begin{eqnarray}
  (\ref{eqn:fdiv_2nd_approx}) & = & 
     {g^{''}(1) \over 2}\epsilon^T \left(\sum_i
     {f^\prime_i {f^\prime_i}^T \over f_i}\right)\epsilon
     \nonumber
     + \\
     \nonumber
     & & {g^\prime (1)\over 2}\epsilon^T\left(
     \sum_if^{''}_i
     \right)\epsilon\\
     \nonumber
     & = & {g^{''}(1) \over 2}
     \sum_i
     {\epsilon^T f^\prime_i {f^\prime_i}^T \epsilon \over f_i}\\
     & = & 
     {g^{''}(1) \over 2}
     \epsilon^T J^T\diag{1\over f} J \epsilon,
     \label{eqn:fdiv_approx_final}
\end{eqnarray}
where $J$ is the input-output Jacobian of $f$, and $\diag{1\over f}$ is a diagonal matrix with its diagonal elements equal to the vector ${1\over f}$. 

}

\section{Model}
\label{sec:model}
Given its superior performance over a wide range of NLP tasks, we focus on exploring different training techniques using BERT \cite{devlin2018bert}. 
We first describe the BERT representations used for all tasks considered in this paper. 
Then, two variants of task-specific BERT-based models are introduced: 
1) the sentence-level classifier for textual entailment and sentiment analysis,
and 2) the extractive QA model.
Specifically, we focus on different ways of encoding input text and building task-specific layers using BERT representations.

\noindent\textbf{BERT Representation:}
For all tasks considered in this work, an input text sequence is divided into subword units $w_t, t=1, \ldots, T$.
The tokenized input sequence is then transformed into embeddings, $\xvec_1, \ldots, \xvec_T\in\RR^n$, through a token encoder, which combines 
a token embedding, a (token) position embedding and a segment embedding (i.e., which text span the token belongs to) by element-wise summation.
The embedding layer is used as the input to multiple transformer layers \cite{vaswani2017attention} to generate the contextual representations, $\hvec_1, \ldots, \hvec_T\in\RR^d$, which are the hidden states of the last layer of the BERT model.
For all regularizations, we sample noise vectors $\epsilon_1, \ldots, \epsilon_T$ from $\mathcal{N}(0, I)$,
and normalize each vector into L2 unit vector.
The noise input is then constructed by adding the normalized noise vector to the token embeddings, \ie $\xvec_1+c\epsilon_1, \ldots, \xvec_T+c\epsilon_T$.
Here, we fix $c=1\mathrm{e}{-3}$ in this paper.

\noindent\textbf{Sentence-level Classifier:}
Following the standard setup of BERT-based textual entailment model \cite{devlin2018bert}, a pair of premise and hypothesis is converted into an input sequence in the form of "\clstoken \textit{premise} \septoken \textit{hypothesis} \septoken". 
Here, \clstoken is a special token indicating the start of the whole sequence and \septoken is another special token for separating the two sentences.
For sentiment analysis, a single sentence is converted to the form of
"\clstoken \textit{sentence} \septoken". 

For both classification tasks, the task-specific layer only takes the first hidden vector $\hvec_{[CLS]}$ produced by BERT, 
corresponding to the \clstoken token.
Then, the probability of class $k$ is
\begin{align}
\label{eqn:class_obj}
P(k|w_1, \ldots, w_T) \propto W^C_k \hvec_{[CLS]},
\end{align}
where $W^C\in\RR^{m\times d}$ is the learnable parameter, the subscript $k$ indicates the $k$-th row of the matrix, and the bias term is left out for simplicity.
For standard BERT training, the log-likelihood based on \autoref{eqn:class_obj} is used. For regularized models, the regularization term is added to stabilize the class probability change with regard to the input noise.

\noindent\textbf{Extractive QA Model:}
For extractive QA, the probability space outcomes consist of token positions of answer spans.
Given a pair of question $q$ and a passage $p$ in the form of "\clstoken \textit{question} \septoken \textit{passage} \septoken",
the BERT encoder produces contextualized representations for all tokens in the input.
Specifically, for each token position $t$ in $p$,
the final hidden vector $\hvec_t \in \RR^d$ 
is used as the contextualized token embedding, where $d$ is the vector dimension.

The span-begin score is computed as 
$s_b(i) = \wvec_b^T \hvec_i$ using a weight vector
$\wvec_b \in \RR^d$.
The probability for a start position $i$ is
\begin{align}
    \label{eqn:span_begin_prob}
    P_b(i) = {\exp(s_b(i)) \over Z_b},
\end{align}
where $Z_b$ is the normalizing factor computed by normalizing over 
$\mathcal{I}$ (the set of all possible positions in the passage), \ie
$Z_b = \sum_{i \in \mathcal{I}} \exp (s_b(i))$.
The span-end score $s_e(j)$, the probability $P_e(j)$ for an end position $j$,
and the normalizing factor $Z_e$ are defined in the same way.
The probability of an answer span $(i, j)$ is
\begin{align*}
\label{eqn:qa_obj}
P(i,j) = P_b(i) P_e(j) = {\exp (s_b(i) + s_e(j)) \over Z_b Z_e}.
\end{align*}
Maximizing the log-likelihood of the above equation is equivalent to maximizing the log probabilities for the correct start and end position, respectively.
For regularized models, given it is computationally expensive to enumerate all possible spans, we apply two separate regularization terms for the start and end position probabilities, respectively.


\section{Experiments}
\label{sec:exp}

In this section, we apply the regularization methods discussed so far to BERT and evaluate their performance on the model robustness.
Specifically, we consider two types of posterior regularization with $f$-divergences.
In addition to a VAT-like regularization with an inner search for the most adversarial direction, following \cite{miyato2018virtual}, we also evaluate the random perturbation training (RPT) with the family of $f$-divergences which only uses randomly sampled noise for regularization. In this work, we focus on three representative $f$-divergences, \ie KL, \textit{squared Hellinger} (SHL) and \textit{Jensen-Shannon divergence} (JSD).

\begin{table*}[h]
    \centering
    \begin{tabular}{l|c|c|c|c}
    \toprule
        Task               &  Training   & Metrics & \multicolumn{2}{c}{Evaluation}\\
                           &             &         & Domain Shift  &  Adversarial Attack\\ \midrule
        Question Answering                 & SQuAD v1.1    & F1/Exact Match (EM) & N/A         &  Adversarial SQuAD \\
        Question Answering                 & SQuAD v2.0    & F1/Exact Match (EM) & BioASQ  & N/A \\
        Textual Entailment & MNLI  & Accuracy (Acc)     & N/A            & HANS \\
        Sentiment Analysis & SST-2   & Accuracy (Acc)     & IMDB & N/A \\
    \bottomrule
        
    \end{tabular}
    \caption{Summary of datasets and their corresponding evaluation purpose and metrics. N/A indicates the trained model is not evaluated for the corresponding generalization. For full data statistics, see \autoref{sec:data_stats}.}
    \label{tab:data_info}
\end{table*}

\noindent\textbf{Dataset:} 
All the datasets used in this paper are summarized in Table~\ref{tab:data_info}.
We consider three tasks, \ie QA, textual entailment, and sentiment analysis, where the last two are sentence classification tasks.
Following the literature, we report the exact match (EM) and F1 scores for QA datasets and classification accuracy for textual entailment and sentiment analysis.
For model training, we use 
MNLI~\cite{mnli2018} and SST-2~\cite{sst2013} 
and SQuAD v1.1/v2.0 \cite{squad1,squad2}, respectively.
The corresponding development set is used for evaluating the in-domain generalization.

To evaluate the out-of-domain generalization with domain shift,
we use the BioAQS dataset \cite{bioasq} from MRQA \cite{mrqa2019} and the IMDB dataset \cite{imdb2011}.
Unlike SQuAD which is based on Wikipedia, BioAQS is a biomedical QA dataset constructed on PubMed articles.
Compared with SST-2 containing pithy export reviews \cite{sst2013}, IMDB includes lengthy movie reviews from non-experts \cite{imdb2011}. 
We directly apply the QA model trained on SQuAD v2.0 and the sentiment classifier trained on SSS-2 to BioAQS and IMDB, respectively.  

To evaluate the model robustness towards adversarial attack, we use two challenging adversarial datasets,
\ie Adversarial SQuAD~\cite{jia2017advsquad} and HANS \cite{mccoy2019acl}
for evaluating QA model trained on SQuAD v1.1 and the textual entailment model trained on MNLI, respectively. 
The Adversarial SQuAD is constructed based on SQuAD v1.1 \cite{squad1} by adding distracting sentences that have high overlap with the question and contain plausible answer candidates.
Naively trained models tend to exploit the word overlap with the given question and thus are fooled by those distracting sentences \cite{jia2017advsquad}.
The HANS dataset is built using three heuristics to ensure that the hypothesis sentence only contains words from the premise sentence \cite{mccoy2019acl}.
Similarly, standard training results in models failing catastrophically, even for BERT.
 
\noindent\textbf{Implementation:}
We follow the default setting used for fine-tuning the uncased BERT base model \cite{devlin2018bert}. 
We select the learning rate from $\{3\mathrm{e}{-5}, 4\mathrm{e}{-5}\}$ for QA models and $\{2\mathrm{e}{-5}, 3\mathrm{e}{-5}\}$ for classification models.
For both tasks, we tune the number of training epochs in $\{2, 3, 4, 5\}$.
In addition, we search regularization weight in $\{0.001, 0.01, 0.1\}$ for JR, and $\{1, 4, 10\}$ for VAT and RPT. 
We use the in-domain dev set for validation and select the best model based on F1 for QA tasks and accuracy for classification tasks.

\begin{table}[t]
\centering
\begin{tabular}{@{\hskip2pt}l|@{\hskip2pt}c|@{\hskip2pt}c|@{\hskip2pt}c|@{\hskip2pt}c|@{\hskip2pt}c}
\toprule
\multirow{2}{*}{\bf Method} 
& v1.1 & v2.0
& {MNLI}  & {SST-2} & Avg \\ 
& F1 & F1 & {Acc} & {Acc} & $\Delta$\\
\midrule

BERT\textsubscript{base} & 88.5 &76.5  & 84.5 & 92.3 & 0.0\\
\midrule
JR  & 88.5 & 74.2  & 84.5 & 92.0 & \colorbox{yellow}{-0.7}\\
\midrule
RPT\textsubscript{KL} & 89.3 & 77.6 & 84.8 & 93.0 & \colorbox{YellowGreen}{+0.7}\\ 
RPT\textsubscript{SHL} & 89.7 & 77.6 & 85.2 & 93.0 & \colorbox{YellowGreen}{+0.9}\\ 
RPT\textsubscript{JSD} & 89.8 & 77.5 & 85.4 & 93.5 &
\colorbox{YellowGreen}{+1.1}\\ 
\midrule
VAT\textsubscript{KL} & 89.9 & 78.9 & 85.3 & 93.1 &
\colorbox{ForestGreen}{+1.4}\\
VAT\textsubscript{SHL} & \textbf{90.2} & 79.2 & 85.5 & 93.2 &
\colorbox{ForestGreen}{+1.6}\\
VAT\textsubscript{JSD} & \textbf{90.2} & \textbf{79.3} & \textbf{85.7} & \textbf{93.6} &
\colorbox{ForestGreen}{\textbf{+1.8}}\\
\midrule
BERT\textsubscript{large} & 90.9 & 81.8 & 86.3 & 93.5 &
\colorbox{ForestGreen}{+2.7}\\
\bottomrule
\end{tabular}
\caption{
Comparison of different training techniques for in-domain generalization on SQuAD (v1.1 and v2.0), MNLI and SST-2 dev sets. BERT\textsubscript{base} and BERT\textsubscript{large} stand for BERT-base and BERT-large model with standard training respectively. The corresponding best performing BERT-base model is in bold. The \texttt{Avg$\Delta$} column is the corresponding average score difference with BERT\textsubscript{base}.
}
\label{tab:in_domain}
\end{table}
\noindent\textbf{In-domain:}
In this part, we focus on comparing the in-domain performance of different training methods.
In other words, each model is trained on the training set and evaluated on the corresponding matched development set.
The experiment is summarized in Table~\ref{tab:in_domain}.
In general, JR performs similarly to the standard BERT training with an exception case for SQuAD v2.0. This is probably because JR uniformly regularizes the Jacobian matrix, which is particularly problematic for QA task with unanswerable questions.
Both RPT and VAT with different $f$-divergences achieve significant improvement over standard training for all four datasets, especially on SQuAD v2.0.
The results suggest incorporating the model confidence into regularization can achieve better in-domain generalization.
Consistent with findings in \cite{miyato2018virtual}, by searching for the most adversarial perturbation direction, VAT variants achieve the largest boost for in-domain generalization.
Moreover, we find that both RPT and VAT with SHL and JSD provides additional improvement over their corresponding counterpart with KL which suggests the benefit of using alternative $f$-divergences for posterior difference regularization. 
Lastly, by selecting the proper divergence, the performance gap between the BERT-base and BERT-large model is dramatically narrowed which indicates the advantage of applying posterior difference regularization with $f$-divergences on top of powerful text representations.

\begin{table}[t]
    \centering
    \begin{tabular}{l|c|c|c}
    \hline
\toprule    
         \multirow{2}{*}{\bf Method} 
          & BioASQ & IMDB & Avg\\
          & F1/EM & Acc & $\Delta$\\ \hline
          BERT\textsubscript{base} &  57.1/41.7	&  87.7 & 0.0\\
          \midrule
          JR &  60.8/46.0 &	87.4 &
          \colorbox{YellowGreen}{+1.7}\\ 
          \midrule
          RPT\textsubscript{KL} &	59.2/43.6 &  \textbf{88.7}
          &\colorbox{YellowGreen}{+1.6}\\
          RPT\textsubscript{SHL} &	60.0/44.8 &  \textbf{88.7}
          &\colorbox{ForestGreen}{+1.9}\\
          RPT\textsubscript{JSD} &	58.3/43.2 &  88.3
          &\colorbox{YellowGreen}{+0.9}\\
          \midrule
          VAT\textsubscript{KL} & 60.1/45.7 & 86.7
          &\colorbox{YellowGreen}{+1.0}\\
          VAT\textsubscript{SHL} & 60.7/45.9 & 87.4
          &\colorbox{YellowGreen}{+1.7}\\
          VAT\textsubscript{JSD} & \textbf{61.8/47.0} & 88.3
          &\colorbox{ForestGreen}{+2.6}\\
          \midrule
          BERT\textsubscript{large} &  63.5/49.5	&  88.3
          &\colorbox{ForestGreen}{+3.5}\\
\bottomrule
    \end{tabular}
    \caption{Domain shift evaluation of different training techniques on BioASQ and IMDB.
    BERT\textsubscript{base} and BERT\textsubscript{large} stand for BERT-base and BERT-large model with standard training respectively.
    The corresponding best performing BERT-base model is in bold.
    The \texttt{Avg$\Delta$} column is the corresponding average score difference with BERT\textsubscript{base}. For BioASQ, F1 score is used for computing the average score.
    EM stands for the exact match score.
    }
    \label{tab:domain_shift}
\end{table}

\begin{table}[t!]
    \centering
\tabcolsep 5pt
    \begin{tabular}{@{\hskip2pt}l|@{\hskip2pt}c|@{\hskip2pt}c|@{\hskip2pt}c@{\hskip2pt}|@{\hskip2pt}c@{\hskip2pt}}
    \hline
\toprule    
         \multirow{2}{*}{\bf Method} 
          & {\small AddSent} & {\small AddOneSent}  & HANS & Avg\\
          & F1/EM & F1/EM & Acc & $\Delta$\\ \midrule
          BERT\textsubscript{base} &  54.0/48.9	& 64.8/59.0 & 58.4
          &0.0\\
          \midrule
          JR &  55.5/49.3 &	65.3/58.4 & 62.6
          &\colorbox{YellowGreen}{+2.1}\\ 
          \midrule
          RPT\textsubscript{KL}&	55.2/49.2 & 66.4/59.8 & 59.9
          &\colorbox{YellowGreen}{+1.4}\\ 
          RPT\textsubscript{SHL}&	57.8/52.2 & 68.9/62.7 & 60.0
          &\colorbox{ForestGreen}{+3.1}\\ 
          RPT\textsubscript{JSD}&	56.0/49.9 & 67.1/61.1 & 61.5
          &\colorbox{YellowGreen}{+2.5}\\ 
          \midrule
          VAT\textsubscript{KL} & \textbf{60.6}/55.0 &	\textbf{69.4}/62.9	& \textbf{67.7}
          &\colorbox{ForestGreen}{+6.8}\\ 
          VAT\textsubscript{SHL} & 60.1/\textbf{55.1} &	69.1/\textbf{63.1}	& 63.2
          &\colorbox{ForestGreen}{+5.1}\\ 
          VAT\textsubscript{JSD} & 58.5/53.2 &	67.4/61.4	& 64.6
          &\colorbox{ForestGreen}{+4.4}\\ 
          \midrule
          BERT\textsubscript{large} & 59.3/54.3 &	69.1/63.6	& 67.9
          &\colorbox{ForestGreen}{+6.4}\\ 
\bottomrule
    \end{tabular}
    \caption{
    Adversarial evaluation of different training techniques on Adversarial SQuAD (AddSent and AddOneSent) and HANS.
    BERT\textsubscript{base} and BERT\textsubscript{large} stand for BERT-base and BERT-large model with standard training respectively.
    The corresponding best performing BERT-base model is in bold. The \texttt{Avg$\Delta$} column is the corresponding average score difference with BERT\textsubscript{base}.
    For Adversarial SQuAD, F1 score is used for computing the average score. EM stands for the exact match score.
    }
    \label{tab:adv_case}
\end{table}
\noindent\textbf{Domain Shift:}
In this part, we compare the performance of models trained using different techniques on datasets with domain shift, \eg different topic or style.
Specifically, we apply the QA models trained on SQuAD v2.0 to the BioAQS version from MRQA \cite{mrqa2019}.
Similarly, we apply the sentiment analysis model trained on SST-2 to the IMDB test set.
The results are summarized in Table~\ref{tab:domain_shift}.
Comparing Table~\ref{tab:domain_shift} with Table~\ref{tab:in_domain},
all methods suffer a noticeable performance drop for both QA and sentiment analysis when evaluated on test sets with domain shift. 
Moreover, we observe more significant performance drop for the QA setting because the biomedical domain differs significantly from the Wiki domain in topic and style, resulting in a larger domain shift between the training and test QA datasets.
Consistent with findings in \cite{hendrycks2020acl}, the in-domain performance is not predictive of the domain shift generalization.
Further, the performance of JR is not stable, with improvement on BioASQ but 
worse performance on IMDB.
Models trained with all three RPT variants result in consistent improvement over standard training on both out-of-domain datasets, suggesting that
random perturbation is particularly effective in enhancing model robustness towards domain shift.
In particular, all RPT variants achieve comparable out-of-domain generalization on IMDB as BERT-Large.
Although all VAT variants achieve decent improvement on BioASQ, neither VAT\textsubscript{KL} nor VAT\textsubscript{SHL} generalize so well to IMDB.
This illustrates the importance of selecting a proper divergence for VAT style regularization. In other words, domain-dependent search for the most adversarial direction with either KL or SHL might be suboptimal for model generalization over domain shift.


\noindent\textbf{Adversarial Attack:}
Here, we evaluate different training techniques on adversarial attack scenarios, where datasets are intentionally constructed to fool naively trained models. 
Specifically, we evaluate 
the QA models trained with SQuAD v1.1 
and the textual entailment models learned on MNLI
using the Adversarial SQuAD and the HANS datasets, respectively.
Table~\ref{tab:adv_case} summarizes the evaluation results of model robustness towards adversarial attacks with different training methods.
For both subsets (AddSent and AddOneSent) from Adversarial SQuAD and HANS, all regularization methods improve over standard BERT training.
In this case, models trained with VAT variants demonstrate stronger resilience towards learning superficial cues from data.
Specifically, VAT with KL achieves the largest improvement on both settings which indicates that an asymmetrical divergence might be more effective in avoiding learning data biases.
Although better text representations derived from BERT-Large are still more robust against adversarial attack than the base version, this gap can be effectively reduced by regularizing the posterior difference with $f$-divergences.
Compared with the recent debiasing method proposed in \cite{clark2019emnlp} that requires the knowledge of existing data bias, VAT variants can be an effective task-agnostic debiasing approach with better in-domain performance and comparable improvement for adversarial settings.

\begin{table}[t!]
    \centering
    \begin{tabular}{l|c|c|c|c}
    \hline
\toprule    
          & SST-2& IMDB & MNLI & HANS\\
          \midrule
          BERT\textsubscript{full} &  92.3 & 87.7 & 84.5 & 58.4\\
          \midrule
          BERT\textsubscript{base} & 91.2 & 86.3 & 82.7 & 51.5 \\
          \midrule
          RPT\textsubscript{KL}&	91.3 & 87.0 & 83.6 & 53.8\\
          RPT\textsubscript{SHL}&	91.9 & 86.6 & 83.8 & 53.3\\
          RPT\textsubscript{JSD}&	91.7 & 86.5 & 83.7 & 51.8\\
          \midrule
          VAT\textsubscript{KL} & 92.1 & 86.3	& 83.1 & 54.3\\ 
          VAT\textsubscript{SHL} & 92.4 & 86.5	& 84.4 & 51.8\\ 
          VAT\textsubscript{JSD} & 92.2 & 86.6	& 84.1 & 52.6\\ 
\bottomrule
    \end{tabular}
    \caption{Comparison of different methods with semi-supervised learning in classification accuracy.
    Except BERT\textsubscript{full}, all others are trained with $50\%$ labelled data.
    }
    \label{tab:semi_half}
\end{table}

\noindent\textbf{Semi-supervised Learning:}
One advantage of regularization methods is their compatibility with semi-supervised learning. 
Given JR is not very effective for the fully-supervised learning, we focus on evaluating RPT and VAT with $f$-divergences under the semi-supervised setting.
Specifically, we use the two sentence classification datasets, MNLI and SST-2, for training.
We hold out $50\%$ of the label information for the training data.
For standard BERT training, only the labelled part is used. For both RPT and VAT variants, the rest unlabelled data is also included for training. Both the cross entropy loss and the regularization term are optimized for the labelled samples, whereas only the regularization term is used for unlabelled ones.
Similar to the fully supervised setting, the models trained on MNLI is applied to HANS for evaluating the model robustness towards adversarial attack, and the models using SST-2 are applied to IMDB to assess the model performance under domain shift.
Results are summarized in Table~\ref{tab:semi_half}.

Compared with the fully supervised setting, all methods get lower classification accuracy across the board.
Both RPT and VAT variants again improve over standard training for both in-domain and out-of-domain evaluations.
It is worth mentioning that both SHL and JSD based VAT models trained with $50\%$ labelled data on SST-2 and MNLI are on par with the corresponding standard BERT training with the full training set which illustrates the advantage of choosing a proper $f$-divergence for the semi-supervised setting.
With only half labelled data, Both RPT and VAT suffer a large drop on HANS and produce almost random predictions, indicating the complimentary benefits of data diversity.
We also further reduce the amount of labelled training data and observe the same trend where regularizing with different $f$-divergences can lead to improved data efficiency. This demonstrates the potential of posterior differential regularization for NLP with low-resource scenarios.

\section{Related Work}
\label{sec:related_work}
With the goal of developing more robust NLP models, a line of recent work has been devoted to identifying various kinds of superficial patterns learned by high-performance models over many popular datasets \cite{gururangan2018annotation,jia2017advsquad,mccoy2019acl}. The prevalence of data \textit{biases} over popular datasets poses a real challenge of accurately estimating the model capacity for practical applications, because a closed dataset evaluation usually inflates the model performance. This concern about dataset biases has led researchers to develop new diagnostic datasets \cite{jia2017advsquad,mccoy2019acl,nie2019adversarial} and training techniques \cite{clark2019emnlp,he2019unlearn} to overcome those discovered biases.
Recent debiasing methods \cite{clark2019emnlp,he2019unlearn} require learning multiple models to access known data biases for the target dataset.
Moreover, they achieve more robust out-of-domain generalization at the price of 
in-domain performance degradation.
In contrast, we focus on the task-agnostic robust learning framework for enhancing model robustness and empirically show that regularization approaches under this framework can result in superior in-domain and out-of-domain generalization.

Training with noise has been a very popular approach for enhancing model robustness.
The dropout is a widely-used approach in deep learning to improve model generalization \cite{dropout}.
For adversarial learning methods, the main theme is reducing the model sensitivity toward 
small input perturbations \cite{goodfellow2014explaining, madry2017pgd},
which has been recently applied to both fine-turning \cite{jiang2019smart,pereira2020alic,zhu2019freelb,li2020textat} and pre-training \cite{liu2020alum}.
However, models trained with adversarial learning are found to have at-odd generalization \cite{tsipras2018robustness,zhang2019theoretically}. 
Our work studies learning methods with the goal of regularizing the model posterior difference of clean and noisy inputs.
We show that compared with the standard BERT training, the proposed posterior differential regularization with $f$-divergence lead to better NLP model robustness.

\section{Conclusion}
\label{sec:conclusion}
\vspace{-2mm}
In this paper, we investigate methods regularizing the posterior difference between the clean and noisy inputs for improving model generalization for both in-domain and out-of-domain settings.
Specifically, we present theoretical analyses of three methods under this framework, \ie 
VAT, JR, and RPT. 
We further extend both VAT and PRT to the family of $f$-divergences and theoretically characterize them in terms of Jacobian matrix.
We also demonstrate their effectiveness in enhancing model robustness over a diverse set of NLP tasks under both fully-supervised and semi-supervised scenarios.

For future work, it is interesting to explore posterior differential regularization methods for weakly-supervised learning, such as relation extraction and QA with distant supervision.

\section*{Acknowledgement}
We would like to thank the anonymous reviewers for valuable suggestions.
Yaoliang Yu thanks NSERC for funding support.

\bibliography{ref}
\bibliographystyle{acl_natbib}
\clearpage
\appendix
\section{Second-order Approximation of Posterior Differential Regularization with $f$-divergence}
\label{apd_sec:approx_reg_derive}

Based on \autoref{eqn:fdiv}, it is easy to show that the first and second derivatives of $D$ with regard to $\epsilon$ are
\begin{eqnarray*}
    D^\prime & = & \sum_i
    g^\prime(\rvec_i)
    f_i^\prime(\hat{\xvec}), \\
    D^{\prime\prime} & = & \sum_i\left[
     g^{\prime\prime}(\rvec_i)   
    {f^{\prime}_i(\hat{\xvec}){f^\prime_i}^T(\hat{\xvec}) \over f_i(\xvec)}
    + 
    g^\prime(\rvec_i)
    f_i^{\prime\prime}(\hat{\xvec})
    \right],
\end{eqnarray*}
where $\rvec_i = {f_i(\hat{\xvec}) \over f_i(\xvec)}$.
Let $D(\epsilon) = D(f(\hat{\xvec}), f(\xvec))$.
When evaluating at $0$, the second order approximation of the $f$-divergence is
\begin{eqnarray}
    D(\epsilon) \approx D(0) + {D^{\prime}}^T(0)\epsilon + {\epsilon^T D^{\prime\prime}(0) \epsilon \over 2}.
    \label{eqn:fdiv_2nd_approx}
\end{eqnarray}
Given that 
\begin{eqnarray*}
D(0) & = & \sum_i g(1)f_i = 0, \\
D^\prime(0) & = & \sum_i g^\prime(1)f_i^\prime = g^\prime(1)
\left(\sum_i f_i\right)^\prime = 0, \\
\sum_i f^{''}_i & = & \left(\sum_i f_i\right)^{''} = 0
\end{eqnarray*}
the second order approximation can be simplified as
\begin{eqnarray}
  (\ref{eqn:fdiv_2nd_approx}) & = & 
     {g^{''}(1) \over 2}\epsilon^T \left(\sum_i
     {f^\prime_i {f^\prime_i}^T \over f_i}\right)\epsilon
     \nonumber
     + \\
     \nonumber
     & & {g^\prime (1)\over 2}\epsilon^T\left(
     \sum_if^{''}_i
     \right)\epsilon\\
     \nonumber
     & = & {g^{''}(1) \over 2}
     \sum_i
     {\epsilon^T f^\prime_i {f^\prime_i}^T \epsilon \over f_i}\\
     & = & 
     {g^{''}(1) \over 2}
     \epsilon^T J^T\diag{1\over f} J \epsilon,
     \label{eqn:fdiv_approx_final}
\end{eqnarray}
where $J$ is the input-output Jacobian of $f$, and $\diag{1\over f}$ is a diagonal matrix with its diagonal elements equal to the vector ${1\over f}$.

\section{Dataset Statistics}
\label{sec:data_stats}

\begin{table}[h!]
	\begin{center}
		\begin{tabular}{@{\hskip2pt}l|c|c|c|@{\hskip2pt}c}
			\toprule 
			\bf Corpus & Train & Dev & Test &Metrics\\
			\midrule
			SST-2 & 67k & 872 & - &  Acc\\
			MNLI & 393k& 20k & -   &   Acc\\
			SQuAD v1 & 87.6k& 10.5k& - & EM/F1\\
			SQuAD v2 & 130.3k &11.9k&-& EM/F1\\
			\midrule
			IMDB   & - & - & 25k &  Acc\\
			BioASQ & - & 1504 & - &  EM/F1 \\
			AddSent & - & - & 3560 & EM/F1\\
			AddOneSent & - & - & 1787 & EM/F1\\
			HANS & - & - & 30k &  Acc\\
			\bottomrule
		\end{tabular}
	\end{center}
	\caption{Experiment dataset summary. The top four datasets are used for training and the matched dev sets are used for both validation and in-domain evaluation. The rest five datasets are used only for evaluation purpose.
	}
	\label{tab:datasets}
\end{table}
The statistics of all datasets used in this paper is summarized in \autoref{tab:datasets}.
The top four datasets in \autoref{tab:datasets} are used for training and the matched dev sets are used for both validation and in-domain evaluation.
The remaining datasets are used only for evaluation purpose. Among the evaluation datasets, other than BioASQ \cite{bioasq} from MRQA where the development set is used for evaluation, the corresponding test set is used for evaluating model out-of-domain generalization.

\end{document}